\documentclass[conference]{IEEEtran}

\usepackage[utf8]{inputenc}
\usepackage[T1]{fontenc}

\IEEEoverridecommandlockouts
\usepackage{amsmath,amssymb,amsfonts}
\usepackage{graphicx}
\usepackage{booktabs}
\usepackage[numbers]{natbib}
\usepackage{colortbl}
\usepackage{multirow}
\usepackage{xspace}
\newcommand{\sys}{{ComInsight}\xspace}

\def\BibTeX{{\rm B\kern-.05em{\sc i\kern-.025em b}\kern-.08em
    T\kern-.1667em\lower.7ex\hbox{E}\kern-.125emX}}
\begin{document}

\title{Structured Composition of Verifiable Atomic Insights for Table-to-Report Generation\\
}

\author{\IEEEauthorblockN{Teng Lin}
\IEEEauthorblockA{\textit{DSA Thrust} \\
\textit{HKUST(GZ)}\\
Guangzhou, China \\
tlin280@connect.hkust-gz.edu.cn}
\and
\IEEEauthorblockN{Xinyu Liu}
\IEEEauthorblockA{\textit{DSA Thrust} \\
\textit{HKUST(GZ)}\\
Guangzhou, China \\
xliu@connect.hkust-gz.edu.cn}
\and
\IEEEauthorblockN{Nan Tang}
\IEEEauthorblockA{\textit{DSA Thrust} \\
\textit{HKUST(GZ)}\\
Guangzhou, China \\
nantang@hkust-gz.edu.cn}

}

\maketitle

\begin{abstract}

Table-to-report generation refers to the task of automatically generating article-level analytical reports from relational tables and is an essential capability for automated data science and decision support. Its central challenge lies in systematically discovering verifiable composite insights across tables, attributes, and analytical perspectives, and organizing them into coherent, complete, and traceable evidence chains. Existing methods primarily rely on sequential, reactive data agents or direct Large Language Model(LLM) generation. They suffer from exploration bias: early local observations constrain subsequent actions, causing models to focus prematurely on local analyzes and miss cross-table or cross-dimensional evidence. We propose \sys, which reformulates insight discovery as the composition of atomic evidences. We first define an atomic insight as the smallest executable analytical unit conforming to a predefined analysis pattern and enumerate all valid atomic insights from database schema and content. These atoms are then organized into a multi-relational insight graph, where nodes represent verified data facts and edges encode logical, temporal, or hierarchical relations. Finally, a set of composition operators systematically fuses atomic nodes into higher-order composite conclusions. Every composite output is accompanied by executable SQL and fine-grained provenance, ensuring full verifiability. Across three benchmarks InsightBench, DDR-Bench, and T2R-Bench, \sys consistently outperforms strong baselines in factual correctness, novelty, and structural completeness. We believe \sys offers a reliable, efficient, and explainable path toward table-to-report generation.

\end{abstract}

\section{Introduction}

\begin{figure*}[h]
  \centering
 \includegraphics[width=\linewidth]{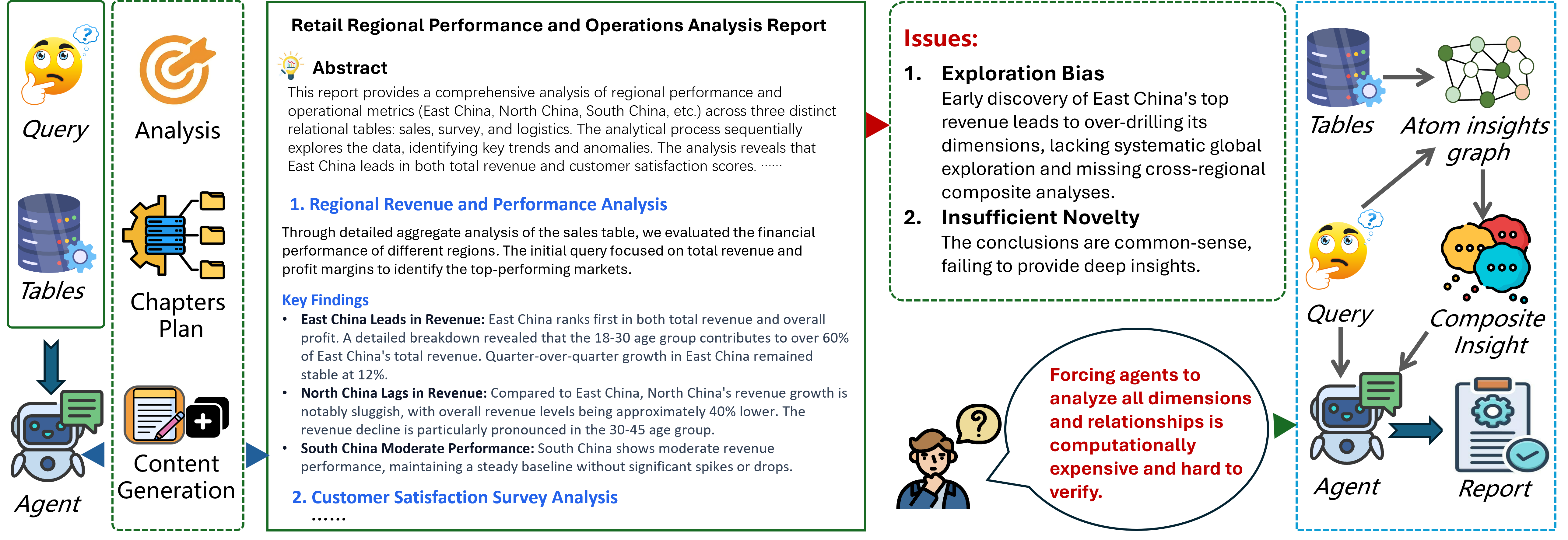}
  \caption{Existing agent workflows suffer from exploration bias and lack of novelty. Exhaustive dimension analysis is unverifiable. The \sys framework shifts from sequential exploration to structured composition, using atom insight graphs to generate verifiable composite insights.}
  \label{fig:intro}
\end{figure*}
Table-to-report generation is the task of automatically synthesizing article-level analytical reports from relational tables, and it constitutes a fundamental capability for automated data science and decision support~\cite{zhang2025t2rbench,lin2026montecarlotree}. Its central challenge lies not only in generating fluent text but in systematically discovering valuable insights across tables, attributes, and analytical perspectives, and organizing them into coherent, complete, and traceable evidence chains~\cite{liu2026huntinstead, huang2026mmdeepresearch,lin2025structured, lin2026eidosdoc}.

An individual finding often captures only one aspect of the data, whereas a comprehensive analysis requires connecting multiple findings to reveal relationships that are not apparent from any single finding alone. We refer to such higher-order findings, supported by multiple atomic findings and explicit relationships among them, as \textbf{composite insights}.
Consider a retail analytics scenario as illustrated in Figure~\ref{fig:intro}. The database contains three tables: sales (order\_id, region, product\_category, age\_group, quarter, revenue, profit), survey (survey\_id, region, quarter, satisfaction\_score, nps\_score), and logistics (shipment\_id, region, quarter, delivery\_days, delay\_flag). A valid analysis should produce a composite insight such as: \emph{``Driven by the 18-30 age group, East China leads in both revenue and satisfaction; in contrast, North China exhibits similar satisfaction but lags significantly in revenue growth, likely because its logistics delay rate (38\% vs. East China's 12\%) constrains revenue conversion.''} This conclusion integrates five independent evidence chains drawn from three different tables and involves four distinct analysis patterns (MAX/GROUP\_BY, AVG/GROUP\_BY, GROUP\_BY with filter, and CORR).

Yet existing data agents struggle to produce such insights~\cite{liu2026huntinstead,sahu2025insightbench,lin-etal-2025-mebench}. Sequential exploration agents, such as ReAct-style agents(e.g., AgentPoirot)~\cite{sahu2025insightbench}, after discovering that ``East China has the highest revenue'' in early interactions, tend to deepen along this direction, repeatedly querying further details about East China, rather than proactively crossing over to the survey table to examine the satisfaction dimension, let alone further linking to the logistics table's delay rate. DDR-Bench's failure analysis shows that 58\% of errors stem from insufficient exploration breadth rather than single-step execution errors~\cite{liu2026huntinstead}. 
These failures expose a fundamental limitation: current data agents lack the capability for systematic discovery of composite insights. 
To address this problem, we propose \sys, an evidence-driven framework for composite insight discovery in table-to-report generation. \sys follows three steps: it first systematically enumerates all valid atomic insights from the database according to a predefined analysis grammar; it then organizes these insights into a typed evidence graph, where nodes represent verified data facts and edges encode relations; finally, relation-specific composition operators fuse independent atomic insights into higher-order composite conclusions. Every composite output is accompanied by executable SQL, ensuring full verifiability.
Our main contributions are as follows:

\begin{itemize}
\item \textbf{We formalize the problem of composite insight discovery for table-to-report generation} and propose five structural composite types: \textbf{comparative, correlative, contextual, temporal, and causal}. Each structural composite type is equipped with explicit composition conditions and reasoning semantics.\item \textbf{We propose the \sys framework}, comprising three core modules: atomic insight generation, multi-relational graph construction, and composite insight generation, along with a Query Sematic Filter that tailors outputs to user-specific analytical goals.\item \textbf{We conduct a systematic evaluation on three benchmarks}: InsightBench, DDR-Bench, and T2R-Bench. Experimental results show that \sys significantly outperforms existing baselines in factual correctness, insight novelty, and structural completeness. 
\end{itemize}

\section{Related Work}

\paragraph{Table Understanding and Reasoning Benchmarks.}
Early TableQA benchmarks such as WikiTableQuestions~\cite{WikiTableQue}, HybridQA~\cite{chen-etal-2020-hybridqa}, FeTaQA~\cite{nan2022fetaqa}, and TAT-QA~\cite{zhu-etal-2021-tat} laid the foundation for question answering over tabular data. More recent comprehensive benchmarks include TableBench~\cite{wu2025tablebench}, MMTU~\cite{xing2026mmtu}, SciTableQA~\cite{ajayi2025scitableqa}, SciNumQA~\cite{son2026scinumqa}, TopBench~\cite{ji2026topbench}, and M3TQA~\cite{zha2024m3tqa}, which extend to complex layouts, multi-table reasoning, multilingual settings, and predictive queries. For table-to-text generation, ToTTo~\cite{parikh2020totto} and FreshTab~\cite{onderkova2025freshtab} are representative efforts. Despite this diversity, all these benchmarks focus on atomic tasks, answering a single question or generating a single description, and do not address the combination of multiple findings into higher-level narratives.

\paragraph{Data Analysis Agents and Insight Generation.}
LLM-based agents have spurred benchmarks for end-to-end data analysis requiring information structuring~\cite{lin2025srag,lin2025structured,lin2025lightkgg,lin2026docsage,lin2026structurethenquery},multi-step reasoning and tool use~\cite{lin2025Simplifying,lin2026annoretrieve}. InfiAgent-DABench~\cite{hu2024infiagent}, DataSciBench~\cite{zhang2025datascibench}, UniDataBench~\cite{weng-etal-2026-unidatabench}, and WebDS~\cite{hsu2025webds}. For insight generation, InsightBench~\cite{sahu2025insightbench}, InsightEval~\cite{zhu-etal-2026-insighteval} and MedInsightBench~\cite{zhu2025medinsightbench} evaluate agents on producing analytical insights, while DDR-Bench~\cite{liu2026huntinstead}, MMDR-Bench~\cite{huang2026mmdeepresearch}, and T2R-Bench~\cite{zhang2025t2rbench} target open-ended deep research and article-level report generation. Evaluation has shifted from moved beyond traditional metrics (e.g., BLEU, ROUGE, and BERTScore) toward LLM-as-a-Judge. Systems such as DataShot~\cite{wang2019datashot}, InsightPilot~\cite{ma2023insightpilot}, InsightLens~\cite{weng2024insightlens}, AgentPoirot~\cite{sahu2025insightbench}, ReActInsight~\cite{weng-etal-2026-unidatabench}, DataCross~\cite{qi2026datacross}, and DABstep~\cite{egg2025dabstep} demonstrate strong exploration and single-insight discovery, but none explicitly composes high-order insights into analytical reports.

\section{Preliminaries and Problem Formulation}

\paragraph{Table-to-Report Task.}We consider a relational database \(\mathcal{D}\) consisting of a set of tables \(\mathcal{T} = \{T_1, \ldots, T_n\}\), where tables are potentially linked through primary-foreign key or semantic associations. Given \(\mathcal{D}\), the task of table-to-report generation is to automatically produce an article-level analytical report $\mathcal{R}$ that summarizes the most important, non-redundant, and verifiable findings in the database. The user query (or report topic) $q$ is optional depending on the specific task. The core problem of table-to-report generation is: \emph{how to define and achieve a comprehensive investigation process that enables a model to discover, compose, and report as many valuable insights in the database as possible, while ensuring traceability, factual correctness, and non-redundancy for each insight?}

\paragraph{Atomic Insight.}An atomic insight is defined as an nine-tuple
\[
a=(id, src, type, dims, metrics, metadata, sql, result, description),
\]
where $type \in \mathcal{T}=\{\mathrm{MAX},\mathrm{MIN},\mathrm{AVG},\mathrm{SUM},\mathrm{COUNT},\mathrm{GROUP\_BY},\mathrm{CORR},\mathrm{TREND}\}$, $\mathcal{T}$ is the set of pattern types. Pattern type is a parameterizable data query template that encapsulates a single data exploration logic.The Patterns and SQL Templates are in Appendix Table~\ref{tab:patterns}. $dims$ and $metrics$ specify the analytical context; $sql$ is the executable query; $result$ is the returned value; and \(description\) is a natural language statement of the finding.

\paragraph{Composite Insight.}An composite insight $c$ is a six-tuple
\[
c=(id,type,atoms,summary,confidence,evidence),
\]
where $atoms$ is the set of constituent atomic insights with $|atoms|\ge 2$, $type\in\{\text{Comparative},$ $\text{Correlative},\text{Contextual},$
$\text{Temporal},\text{Causal}\}$, $summary$ is a natural language statement, $evidence$ contains the SQL queries and results, and $confidence$ measures its credibility in terms of statistical significance and logical consistency. 
Formally, let $\mathcal{A}$ be the set of all possible atomic insights , and $\mathcal{C}$ be the set of all valid composite insights. 

\paragraph{Research Goals.}
The objective is to construct a report $\mathcal{R}$ such that: \textbf{Traceability:} every insight contains a complete evidence chain; \textbf{Factual Correctness:} every insight is supported by verifiable atomic insights and executable SQL; \textbf{Non-Redundancy:} $\mathcal{R}$ does not contain duplicate or semantically equivalent insights. 
\section{The \sys Framework}
\label{sec:framework}


\subsection{Atomic Insight Representation and Generation}
\label{subsec:atom}
\begin{figure*}[h]
  \centering
 \includegraphics[width=\linewidth]{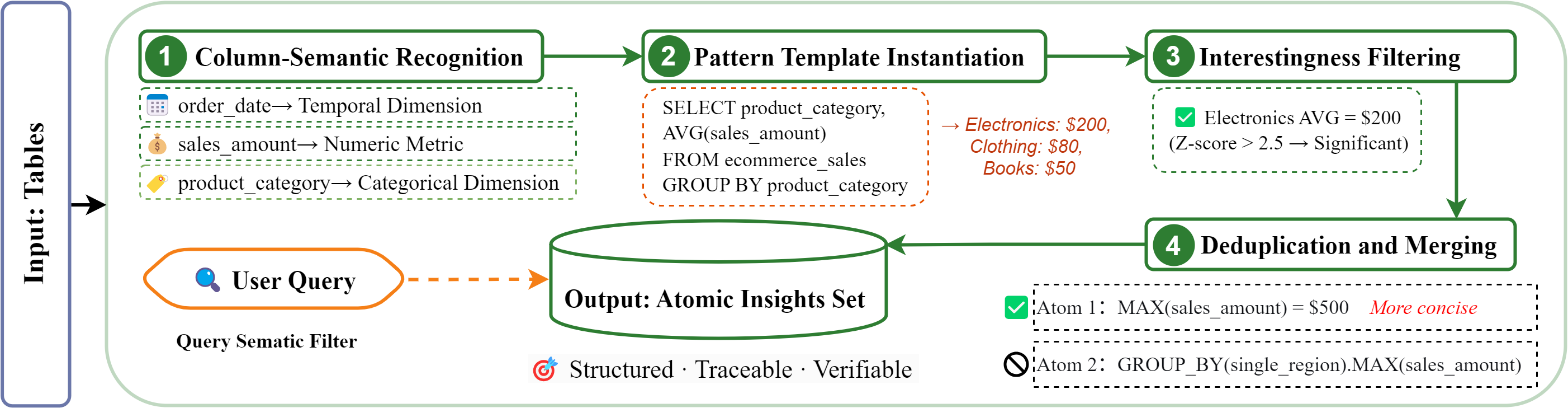}
  \caption{The pipeline of Atomic Insight Generation.}
  \label{fig:atom}
\end{figure*}

The generation pipeline is illustrated in Figure~\ref{fig:atom}. Atomic insights are generated through \textbf{Column Semantic Recognition, Pattern template Instantiation, Interestingness Filter}(statistical significance filter and LLM-based semantic filter), and \textbf{Semantic Deduplication}, yielding a non-redundant set $\mathcal{A}$ of atomic insight units. If the query or topic is given, we apply a \textbf{Query Sematic Filter} based on embedding relevance to select top-50 atomic insights before Interestingness Filter. The detailed definition and implementation steps of atomic insights are provided in the Appendix~\ref{sec:atomic}.

\paragraph{Training Data Construction.}
\label{subsec:traindata}
We build training data by selecting a subset of instances from three benchmarks: 40 datasets from InsightBench (out of 100 instances), 100 tasks from DDR-Bench (out of 291), and 150 industrial tables from T2R-Bench (out of 457), totaling 290 instances. These instances serve two purposes: (i) calibrating the statistical significance filtering rules and thresholds, and (ii) training the lightweight LLM-as-Judge. Experts then annotate their semantic interestingness, yielding 1,200 (labeled interesting / not interesting)  samples for supervised fine-tuning (SFT) and 600 preference pairs (Triplet (inputs, chosen, rejected)) for direct preference optimization (DPO).

\paragraph{Statistical significance rules and Interestingness Judge.} 
We first classify training insights into eight pattern types, parse their SQL to extract metrics, dimensions, aggregations, and columns, then select the corresponding statistical test for each type and calibrate thresholds on a validation set by F1, yielding the final significance rules.(e.g., for CORR, retain candidates whose absolute Pearson correlation coefficient $|r|$ exceeds 0.6 and whose significance level $p$ is below 0.05), and rules for the eight types are in Appendix~\ref{subsec:rules}. A lightweight LLM-as-Judge (Llama-3-8B), trained via SFT on labeled samples and DPO on 600 preference pairs, evaluates semantic interestingness. In our experiments, the Interestingness Filter module filters about \textbf{85\%} of atomic insights while preserving high-value discoveries.

\subsection{Multi-Relational Graph Construction}
\begin{figure*}[h]
  \centering
 \includegraphics[width=\linewidth]{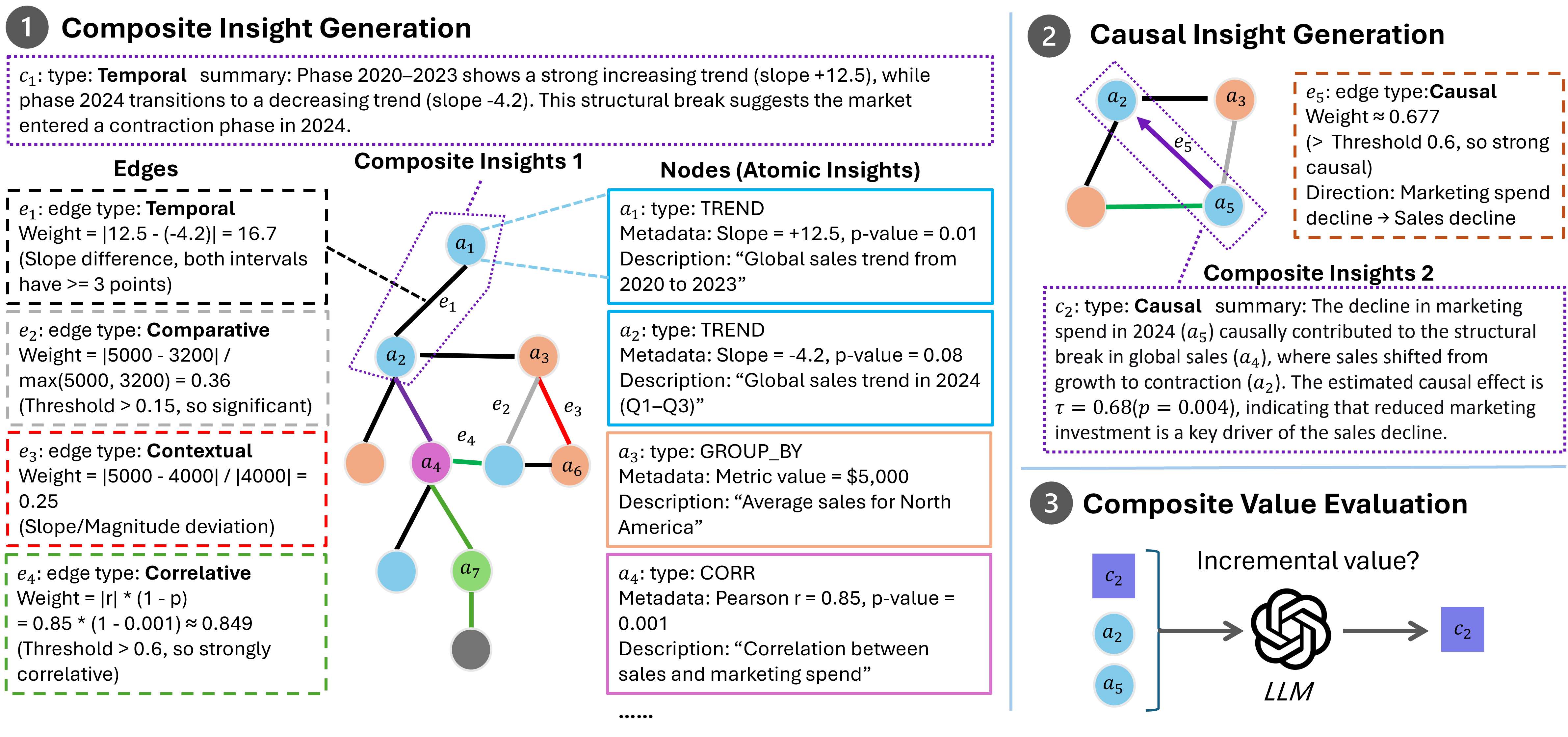}
  \caption{1. Diagram of the multi‑relational graph constructed. Each node represents an atomic insight. Edges between nodes are established according to five relation types. 2. Diagram of Causal insights generation. 3. Composite Value Evaluator.}
  \label{fig:graph}
\end{figure*}

Atomic insights are modeled as nodes in a \textbf{multi-relational graph} $\mathcal{G}=(\mathcal{V},\mathcal{E},\mathcal{R})$ where nodes \(\mathcal{V}\) correspond to atomic insights, \(\mathcal{R} = \{\mathrm{Comparative}, \mathrm{Correlative}, \mathrm{Contextual}, \mathrm{Temporal}, \mathrm{Causal}\}\) denotes the five relation types, and \(\mathcal{E}\) is the set of typed edges. \textbf{The first four relations are treated as undirected edges}. As illustrated in Figure~\ref{fig:graph}-1, each edge is equipped with a type-specific weight that quantifies the strength of the relation. Comparative edges connect same-metric/aggregation atoms with different dimension values, weighted by normalized relative difference and retained if it $> 0.15$. Correlative edges connect statistically dependent or co-moving trend atoms, and retained if the weight $> 0.6$. Contextual edges link global aggregates with local segments, weighted by slope or magnitude deviation with sign reversal weighted double. Temporal edges connect adjacent segmented trends partitioned by PELT, weighted by absolute slope difference with a minimum segment length of three. As illustrated in Figure~\ref{fig:graph}-2, \textbf{Causal edges are directed} and are established only when two conditions are satisfied simultaneously: (1) a \textbf{Granger causality test} on the corresponding time series of the two atomic insights is statistically significant, i.e., the $p$-value is below $0.05$ after appropriate lag selection; (2) an \textbf{LLM-based causal verifier} judges that a plausible causal relation exists, based on the descriptions, SQL results, and domain context, and assigns a mechanism score $s_{\text{mech}}\in[0,1]$ as well as a confounding-risk score $c_{ij}\in[0,1]$.
Temporal precedence $t_i < t_j$ is required as a prerequisite for the Granger test. The weight of a directed causal edge $i\to j$ is
\begin{equation}
w_{\text{causal}}(i,j)
=
\mathbb{I}[t_i<t_j]
\cdot
\mathbb{I}[p_{\text{Granger}}<0.05]
\cdot
|r_{ij}|(1-p_{ij})
\cdot
s_{\text{mech}}(i,j)
\cdot
(1-c_{ij}),
\end{equation}
where $p_{\text{Granger}}$ is the $p$-value from the Granger causality test, and $r_{ij}$, $p_{ij}$ are the adjusted correlation coefficient and its $p$-value. A directed edge $i\to j$ is retained only if $w_{\text{causal}}(i,j)>0.6$. 
This dual requirement ensures that causal edges are both statistically supported and semantically plausible.

The graph is implemented as a MultiGraph, allowing multiple parallel edges between the same node pair to capture different relational facets. The detailed edge weight computation is shown in Appendix~\ref{sec:graph}.

\subsection{Composite Insight Generation}
\paragraph{Community-based Construction.}
We perform community detection on each of the four undirected relation subgraphs. For each relation type, we run the Leiden algorithm on the subgraph to partition atomic insights into communities. Communities with modularity greater than $0.3$ and size at least $3$ are retained. For each retained community, we prompt an LLM to synthesize a higher-order composite insight that integrates all atomic insights in the community, producing a coherent narrative that reflects the underlying relational structure. The LLM receives the descriptions, types, and SQL results of all constituent atoms, and is instructed to generate a summary that captures the shared theme and relational pattern. The steps and prompts are in Appendix~\ref{subsec:community-aggregation}.

\paragraph{Causal Insights Construction.}
For causal relations, we search directed causal paths of length up to $L_{\max}$ (e.g., $3$). Each path is aggregated by an LLM into a causal composite insight that states the causal chain and includes the Granger $p$-value, mechanism score, and confounding risk. Causal composites are treated as hypotheses and are retained only if every causal edge on the path has been validated by both the Granger test and the LLM judge.

After initial Composite Insight Generation, an LLM-based \textbf{Composite Value Evaluator} compares each composite against its constituent atomic insights to determine retain the atoms or composite, which is shown in Figure~\ref{fig:graph}-3. So the final insights set includes generated composite insights and retain atomic insights. More detailed definition and implementation steps of composite insights are provided in the Appendix~\ref{sec:composite}.

\subsection{Report Generation}
Given the set of insights and an optional user query(topic) $q$, the report generation module first removes insights whose embedding similarity to \(q\) falls below a threshold (or keeps all if \(q\) is absent). Then we group related insights into thematic sections and uses an LLM to generate coherent narratives that cover all retained insights with every claim backed by SQL evidence. The final report is assembled from these sections and remains fully traceable and verifiable. Details is in Appendix~\ref{sec:report-generation}.
\section{Experiments}

\subsection{Experimental Setup}

\paragraph{Datasets.}
We evaluate \sys on three benchmarks’s test sets: \textbf{InsightBench}: Consists of 100 datasets with 475 insights. \textbf{DDR-Bench} (Deep Data Research Benchmark): Contains 291 tasks and 2058 insights. \textbf{T2R-Bench}: Contains 910 questions on 457 real-world industrial tables from 19 industry domains. The LLM-as-Judge is trained on the 290 training instances listed in Section~\ref{subsec:atom}. All training instances are excluded from test. Results are reported on held-out test splits.

\paragraph{Baselines.}

\textbf{Pandas Agent}~\cite{Chase_LangChain_2022}: A data science agent optimized for single-turn question answering. Given a dataframe and a question, it generates and executes Python code to produce the answer. \textbf{AgentPoirot}: The baseline data analysis agent proposed in InsightBench, capable of performing end-to-end data analytics by generating Descriptive, Diagnostic, Predictive, and Prescriptive questions. \textbf{DeepSeek-R1}~\cite{guo2025deepseek}: The best-performing model on T2R-Bench, achieving an overall score of 62.71\%. \textbf{Claude 4.5 Sonnet}~\cite{anthropic2025claudesonnet45}: The best-performing proprietary model on DDR-Bench and \textbf{GPT-5.5}.

\paragraph{Evaluation Metrics.}

We adopt benchmark-specific metrics: For InsightBench, LLaMA-3-Eval (a G-Eval variant based on LLaMA-3-70b) for insight-level and summary-level scores. For DDR-Bench, Checklist-based accuracy, where GPT-5-mini judges whether generated insights support each checklist fact. For T2R-Bench, there are three-dimensional evaluation: Numerical Accuracy Criterion (NAC), Information Coverage Criterion (ICC), and General Evaluation Criterion (GEC), and report their average score.


\subsubsection{Implementation Details}

\sys's atomic insight generation module is based on predefined analysis pattern templates. The multi-relational graph is implemented using NetworkX's MultiGraph structure. 
All experiments are conducted on 8 NVIDIA RTX4090 GPUs. The LLM used for composite insights generation and report generation is gpt-5.5 (OpenAI API). Sampling temperature is set to 0.0 for reproducibility, with each experiment repeated 5 times reporting average results.


\begin{table*}[t!]
\centering
\caption{Performance comparison on InsightBench. I stands for Insight-level, S is Summary-level, and Eval is LLaMA-3-Eval. \textbf{Bold} indicates the best metric,
\underline{Underline} indicates the second best.}
\label{tab:insightbench}
\begin{tabular}{lcccc}
\toprule
Method & I-Eval & S-Eval & I-ROUGE-1 & S-ROUGE-1 \\
\midrule
Pandas Agent (gpt-5.5)
& 0.54 $\pm$ 0.01
& 0.40 $\pm$ 0.04
& \underline{0.35 $\pm$ 0.03}
& 0.35 $\pm$ 0.01 \\
AgentPoirot (gpt-5.5)
& \underline{0.60 $\pm$ 0.03}
& \underline{0.44 $\pm$ 0.03}
& 0.32 $\pm$ 0.02
& \underline{0.37 $\pm$ 0.09} \\
DeepSeek-R1      & 0.47 $\pm$ 0.03 & 0.31 $\pm$ 0.02 & 0.29 $\pm$ 0.02 & 0.31 $\pm$ 0.01 \\
 Claude 4.5 Sonnet  & 0.45 $\pm$ 0.03 & 0.27 $\pm$ 0.02 & 0.23 $\pm$ 0.02 & 0.29 $\pm$ 0.05 \\
GPT-5.5 & 0.51 $\pm$ 0.02 & 0.36 $\pm$ 0.03 & 0.31 $\pm$ 0.02 & 0.33 $\pm$ 0.04 \\
\rowcolor[gray]{0.9}
\textbf{\sys}   & \textbf{0.73 $\pm$ 0.03} & \textbf{0.55 $\pm$ 0.03} & \textbf{0.42 $\pm$ 0.02} & \textbf{0.43 $\pm$ 0.03} \\
\bottomrule
\end{tabular}
\end{table*}

\subsection{Main Results}

\subsubsection{Performance on InsightBench}

Table~\ref{tab:insightbench} presents the performance comparison on InsightBench. \sys significantly outperforms all baselines on both insight-level and summary-level LLaMA-3-Eval scores.
\sys achieves an insight-level LLaMA-3-Eval score of 0.73, outperforming AgentPoirot (gpt-5.5) by 13.3 percentage points. This significant improvement demonstrates that explicit composite insight generation effectively integrates multiple atomic insights, producing richer and more accurate conclusions than single-step analysis. Meanwhile, GPT-5.5 achieves only 0.51, significantly below \sys's 0.73, demonstrating that simple linguistic concatenation cannot replace structured evidence combination reasoning.


\subsubsection{Performance on DDR-Bench}

\begin{figure*}[ht]
  \centering
 \includegraphics[width=\linewidth]{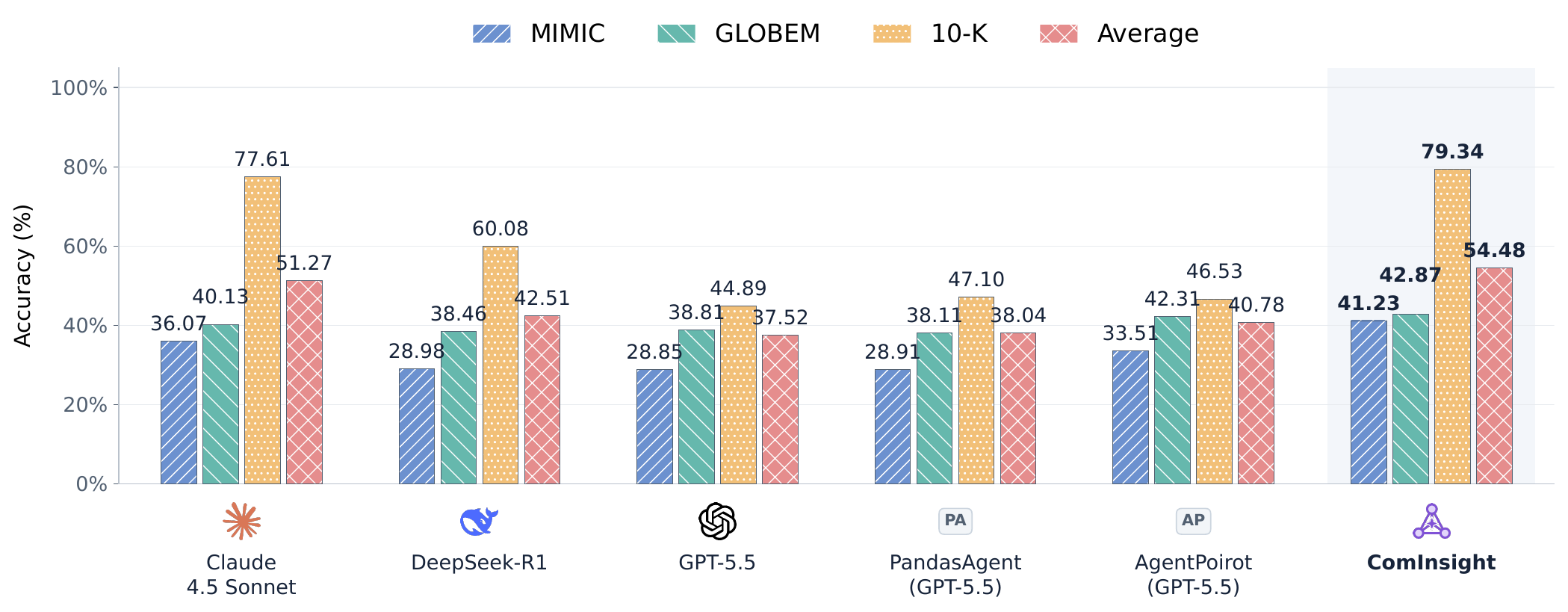}
  \caption{Performance comparison on DDR-Bench.}
  \label{fig:ddrbench}
\end{figure*}

Figure~\ref{fig:ddrbench} presents the performance comparison on DDR-Bench across three scenarios. \sys achieves the best performance on MIMIC, GLOBEM and 10-K scenarios.
On the MIMIC scenario, \sys achieves 41.23\% accuracy, outperforming Claude 4.5 Sonnet by 14.3\%. On the 10-K scenario, \sys reaches 79.34\%, a 2.2\% improvement over the native model. The improvement on GLOBEM is relatively modest (from 42.31\% to 42.87\%), likely because GLOBEM's checklist items are closed-form (three-option multiple choice), leaving less room for composite reasoning.

\begin{figure*}[ht]
  \centering
 \includegraphics[width=\linewidth]{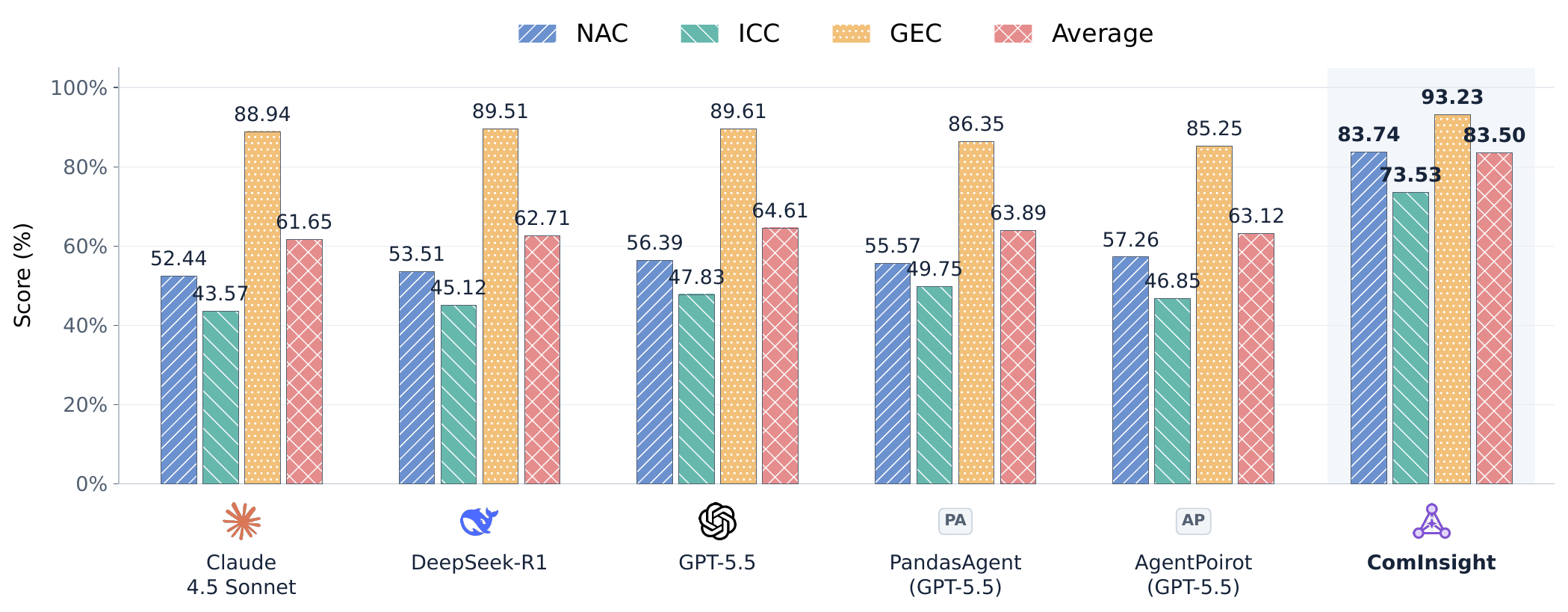}
  \caption{Performance comparison on T2R-Bench.}
  \label{fig:t2rbench}
\end{figure*}



\subsubsection{Performance on T2R-Bench}

Figure~\ref{fig:t2rbench} presents the performance comparison on T2R-Bench. \sys outperforms baselines across all three evaluation dimensions.
\sys achieves an Average score of 83.5\% on T2R-Bench, outperforming native GPT-5.5 (64.61\%) by 18.89 percentage points. The improvement on Numerical Accuracy Criterion (NAC) is most significant (+ 26.48\%), suggesting that explicit evidence chaining in composite insight generation helps reduce numerical hallucinations, the most common error type in T2R-Bench report generation. Information Coverage Criterion (ICC) improves from 49.75\% to 73.73\%, indicating that composite insights more comprehensively cover report keypoints. This finding aligns with \sys's design: through multi-relational graph traversal, the system systematically discovers associations between atomic insights, avoiding omissions of important information.

\subsection{Ablation Studies}

To isolate the contribution of the three key stages in \sys, we conduct ablations on T2R-Bench under the same setting as the main experiment. The evaluated variants are: (1) \textbf{w/o Composite Insight Generation (Simple LLM Concatenation)}, which prompts an LLM(GPT-5.5) to concatenate atomic insights into a report; (2) \textbf{w/o Composite Value Evaluation}, which retains composite candidates but removes the LLM-based incremental-value judge and (3) \textbf{w/o Causal Relationship Generation}, which removes causal assessment and causal composite merging.

\begin{table*}[t]
\centering
\caption{Ablation study on T2R-Bench. $\Delta$Average is computed against full \sys.}
\label{tab:ablation_t2r}
\begin{tabular}{lcc}
\toprule
Configuration & Average(\%) & $\Delta$Average(\%) \\
\midrule
Full \sys & 83.5 & -- \\
w/o Composite Insight Generation (Simple LLM Concatenation) & 64.61 & -18.89 \\
w/o Composite Value Evaluation & 71.34 & -12.16 \\
w/o Causal Relationship Generation & 76.75 & -6.75 \\
\bottomrule
\end{tabular}
\end{table*}

On T2R-Bench, full \sys achieves an Average of 83.5\% ; removing composite insight generation and reverting to simple LLM concatenation reduces Average to 64.61\% ($\Delta$Avg $=-18.89\%$), indicating that structured composition over the multi-relational graph is the dominant source of improvement; removing composite value evaluation yields 71.34\%, showing that the incremental-value judge is important for suppressing redundant or low-utility composites and preserving coverage; and removing causal relationship generation gives 76.75\%, suggesting that causal chains contribute less to T2R-Bench numerical accuracy and information coverage but still support reasoning depth and narrative completeness.

\subsubsection{Efficiency and Time Analysis}
Across the three benchmarks test set, \sys generates on average about 28 atomic insights, 142 typed edges, and 41 final composite insights per task. The end-to-end pipeline takes about 102.3 seconds: 14.2s for atomic enumeration and SQL execution, 17.6s for statistical filtering and semantic evaluation, 1.4s for graph construction, 52.8s for composite generation (51.6\%, the main bottleneck), and 16.3s for report generation. T2R-Bench is the slowest due to larger tables and more atoms, while InsightBench is the fastest; batching LLM calls and parallelizing composite generation can reduce the time by 4–6×.

\section{Conclusion}
In conclusion, this paper addresses the exploration bias and limited novelty of existing data agents in table-to-report generation by formalizing composite insight discovery and proposing \sys. \sys generates verifiable atomic insights, organizes them into a typed multi-relational evidence graph, and applies relation-specific composition operators to fuse them into comparative, correlative, contextual, temporal, and causal composite insights, while query-driven intent alignment and executable SQL provenance ensure relevance, factual correctness, structural completeness, and traceability. Experiments on InsightBench, DDR-Bench, and T2R-Bench demonstrate that \sys substantially outperforms existing baselines in factual correctness, novelty, and structural completeness, and ablation studies confirm the critical roles of the multi-relational graph and specialized composition operators. This work thus offers a principled and verifiable structured-composition paradigm for automated, report-oriented data analysis. 
\section{AI use statement}
 In this work, we used generative AI tools for required-disclosure tasks involved in implementing and evaluating the proposed method: a Llama-3-8B model was fine-tuned via SFT and DPO and used as an LLM-as-Judge for semantic interestingness evaluation; GPT-5.5 was used for composite insight generation and report generation; and LLaMA-3-70B and GPT-5-mini were used as automatic evaluators for benchmark experiments. These uses are part of the \sys methodology and evaluation pipeline and are described in Sections 4 and 5. We have not used generative AI tools for generating synthetic datasets, developing theoretical models or conceptual frameworks, formulating mathematical claims, assisting in the writing of proofs, proposing or refining hypotheses, designing research methodology or experiments, cleaning or reformatting datasets, supporting qualitative/thematic data analysis, or interpreting results beyond the LLM-based components explicitly described above. We have reviewed all AI-assisted work. For example, all LLM-generated outputs used as evidence were checked against executable SQL and experimental results, and all references were verified manually. We take responsibility for the final content of this work, including text, claims, or artifacts produced with the aid of generative AI.

\bibliographystyle{IEEEtran}
\bibliography{ref/custom}

\appendix
\section{Appendices}

\section{The \sys Framework}
\label{sec:framework}

ComInsight takes relational tables as input and produces a set of high-quality composite insights as output. Its design follows a four-stage progression: \emph{atomic discovery} $\rightarrow$ \emph{relational modeling} $\rightarrow$ \emph{structured composition} $\rightarrow$ \emph{causal verification}. The framework consists of three core modules and two auxiliary layers.

\begin{itemize}
    \item \textbf{Atomic Insight Generation Module.} It performs multi-dimensional analysis pattern enumeration over input tables, followed by statistical salience filtering, semantic interestingness evaluation, and semantic deduplication. The output is a set of high-quality atomic insights that serve as elementary evidence units for composition.
    \item \textbf{Multi-Relational Graph Construction Module.} It models atomic insights as nodes and computes type-specific edge weights for four structural relation types: comparative, correlative, contextual, and temporal. The resulting typed evidence graph provides an interpretable search space for composite discovery.
    \item \textbf{Composite Insight Generation Module.} It performs edge-based traversal and type-specific composition, followed by causal verification, deduplication, and confidence scoring. It outputs five categories of composite insights: comparative, correlative, contextual, temporal, and causal.
\end{itemize}

In addition, a \textbf{Query-driven Intent Alignment Layer} sits between atomic generation and graph construction. Given a user query, it filters the atomic insight pool by semantic relevance, so that subsequent composition focuses on directions genuinely relevant to the user. A \textbf{Semantic Interestingness Judge} and a \textbf{Causal Verifier} are integrated into the atomic and composite stages, respectively.

\subsection{Atomic Insight Representation and Generation}
\label{sec:atomic}

\subsubsection{Definition of Atomic Insight}

An atomic insight is the smallest executable analytical unit conforming to a predefined analysis grammar. Formally, an atomic insight $a$ is an nine-tuple:
\begin{equation}
a = (id, src, type, dims, metrics, metadata, sql, result, description),
\end{equation}
where:
\begin{itemize}
    \item $id$: a globally unique identifier;
    \item $src$: the data source, e.g., a database table or a cross-table view;
    \item $type \in \mathcal{T}$: the analysis pattern type, where 
    
    $\mathcal{T} = \{\mathrm{MAX}, \mathrm{MIN}, \mathrm{AVG}, \mathrm{SUM}, \mathrm{COUNT}, \mathrm{GROUP\_BY}, \mathrm{CORR}, \mathrm{TREND}\}$;
    \item $dims$: the set of dimension columns and their filter values, e.g., $\{\mathrm{region}:\mathrm{East}, \mathrm{quarter}:\mathrm{Q1}\}$;
    \item $metrics$: the set of metric columns and their aggregations, e.g., $\{\mathrm{revenue}:\mathrm{SUM}\}$;
    \item $metadata$: structured numerical evidence, e.g., ``slope $= +12.5$, $p$-value $= 0.01$'';
    \item $sql$: the executable query used to reproduce and verify the insight;
    \item $result$: the numeric or tabular result returned by the SQL query.
\end{itemize}
Let $\mathcal{A}$ denote the space of all atomic insights. However, not every generated candidate is retained. An atomic insight is included in the final evidence pool only if it passes both statistical salience and semantic interestingness checks.

\subsubsection{Generation Pipeline for Atomic Insights}

The atomic insight generation module takes arbitrary relational tables as input and produces a curated set of atomic findings through five stages.

\paragraph{Training Data Construction.}
We build training data by selecting a subset of instances from three benchmarks: 40 datasets from InsightBench (out of 100 instances), 100 tasks from DDR-Bench (out of 291), and 150 industrial tables from T2R-Bench (out of 457), totaling 290 instances. These instances serve two purposes: (i) calibrating the statistical salience filtering rules and thresholds, and (ii) training the lightweight LLM-as-Judge.

Specifically, from these 290 instances we automatically enumerate approximately 3,200 candidate atomic insights. Domain experts then annotate their semantic interestingness, yielding 1,200 supervised fine-tuning (SFT) samples and 600 preference pairs for direct preference optimization (DPO).

\paragraph{Stage 1: Column Semantic Recognition.}
The system classifies each column into three semantic categories:
\begin{itemize}
    \item \textbf{Dimension Column:} low-cardinality categorical columns (cardinality $\leq 50$), or numerical columns with a small value set that semantically serves as grouping criteria, e.g., ``region'';
    \item \textbf{Metric Column:} high-cardinality numerical columns suitable for aggregation, e.g., SUM, AVG, MAX/MIN;
    \item \textbf{Time Column:} datetime/timestamp columns, or columns whose names contain time-related keywords such as ``date'', ``time'', ``year'', ``quarter'', or ``month''.
\end{itemize}
The classification combines column statistics and naming heuristics.

\paragraph{Stage 2: Pattern Template Instantiation.}
For each combination of metric columns and dimension columns, the system instantiates parameterized SQL templates for the analysis patterns in $\mathcal{T}$. This stage enumerates all reasonable dimension--metric--pattern combinations and yields a set of raw candidate atomic insights.

\paragraph{Stage 3: Statistical Significance Filtering.}
\label{subsec:rules}
We calibrate the significance thresholds for each analysis pattern by processing the insights in the training dataset as follows:
\begin{enumerate}
    \item \textbf{Type classification}: Classify each insight in the training set into one of eight types according to its analysis pattern: MAX, MIN, AVG, SUM, COUNT, GROUP\_BY, CORR, TREND.
    \item \textbf{SQL parsing}: Parse the SQL query of each insight to extract the metric columns, dimension columns, aggregation functions, grouping conditions, time columns, returned values, and metadata.
    \item \textbf{Statistical test selection}: Choose the corresponding statistical test based on the type:
    \begin{itemize}
        \item CORR: compute the Pearson correlation coefficient \(r\) and its significance \(p\)-value;
        \item TREND: fit a linear regression of the metric over time to obtain the slope and its significance \(p\)-value;
        \item GROUP\_BY: compute the standardized deviation of each group aggregate from the global mean;
        \item MAX/MIN: identify extrema;
        \item AVG, SUM, COUNT: compute distributional statistics for subsequent thresholding.
    \end{itemize}
    \item \textbf{Threshold determination}: On a validation set, select the threshold that optimizes F1 or agreement with expert judgments. The final rules are:
    \begin{itemize}
    \item \textbf{MAX / MIN}: Extrema are inherently interesting; retain all extrema without additional thresholds, i.e., \(S_{\mathrm{stat}}(a)=1\).
    \item \textbf{AVG}: If it is a global average, retain it directly. If grouped by a dimension, compute the standardized deviation of each group mean from the global mean, \(z = |\bar{x}_g - \mu| / (\sigma / \sqrt{n})\), and retain groups with \(|z| > 2\).
    \item \textbf{SUM}: If it is a global sum, retain it directly. If grouped, compute the standardized deviation of each group sum from the global sum, or its proportion, and retain groups whose deviation exceeds two standard errors or whose proportion exceeds the 90th percentile.
    \item \textbf{COUNT}: If it is a global count, retain it directly. If grouped, use a binomial test or chi-square test and retain groups with \(p < 0.05\).
    \item \textbf{GROUP\_BY}: Retain groups whose aggregate deviates from the global mean by more than two standard errors, with multiple-comparison correction.
    \item \textbf{CORR}: Retain candidates with \(|r| > 0.6\) and \(p < 0.05\).
    \item \textbf{TREND}: Retain candidates with regression \(p < 0.05\).
\end{itemize}
\end{enumerate}
We define a binary statistical salience indicator \(S_{\mathrm{stat}}(a) \in \{0,1\}\), where \(S_{\mathrm{stat}}(a)=1\) if atomic insight \(a\) passes the rule for its corresponding type, and \(0\) otherwise.

\paragraph{Stage 4: Semantic Interestingness Evaluation.}
Statistical significance alone does not guarantee that an atomic insight is semantically interesting, novel, or actionable. We therefore train a lightweight \textbf{LLM-as-Judge} to score the semantic interestingness of each candidate.

We use Llama-3-8B as the judge. The training data come from the 290 instances described above: candidate atomic insights are generated from these instances, and experts annotate their interestingness, forming 1,200 SFT samples and 600 DPO preference pairs.

\textbf{Base Configuration and LoRA Setup.}
We adopt QLoRA for parameter-efficient fine-tuning with 4-bit NF4 quantization and bf16 computation. The LoRA configuration is: rank \(r=16\), \(\alpha=16\), dropout \(=0.05\), with target modules \texttt{q\_proj}, \texttt{k\_proj}, \texttt{v\_proj}, \texttt{o\_proj}. The optimizer is AdamW with weight decay \(0.05\), and the learning rate schedule is cosine decay with a warmup ratio of \(0.1\). All experiments are conducted on 8 NVIDIA RTX 4090 GPUs with sampling temperature set to \(0.0\) for reproducibility.

\textbf{SFT Stage.}
We perform supervised fine-tuning on 1,200 expert-annotated samples for 3 epochs. The learning rate is \(2\times 10^{-5}\), per-device batch size is \(4\), gradient accumulation steps is \(4\), yielding an effective batch size of \(128\). The maximum sequence length is 2,048 tokens. Loss is computed only on response tokens, not on prompts.

\textbf{DPO Stage.}
Starting from the SFT model, we apply DPO on 600 preference pairs for 1 epoch. The learning rate is \(5\times 10^{-7}\), per-device batch size is \(2\), gradient accumulation steps is \(8\), yielding an effective batch size of \(16\). The DPO temperature \(\beta\) is set to \(0.1\), and the maximum sequence length is 2,048 tokens. The number of warmup steps is \(100\). The training data are derived from 290 benchmark instances (40 from InsightBench, 100 from DDR-Bench, and 150 from T2R-Bench).


\paragraph{Stage 5: Deduplication and Merging.}
Different analysis patterns may produce semantically equivalent or near-redundant insights, e.g., a MAX result and a GROUP\_BY extreme referring to the same fact. We normalize each atomic insight by the triple $(\mathrm{ description}, \mathrm{metrics}, \mathrm{type})$ and merge equivalent items, keeping the one with the most concise $description$. 

\subsubsection{Query Sematic Filter}

Given a user query $q$ and the atomic insight set $\mathcal{A}'$, the intent alignment layer computes a relevance score
\begin{equation}
r(q,a) = \mathrm{sim}\big(\mathrm{enc}(q), \mathrm{enc}(a.\mathrm{, description} \oplus a.\mathrm{metrics} \oplus a.\mathrm{type})\big),
\end{equation}
where $\mathrm{enc}(\cdot)$ is a text embedding model, $\oplus$ denotes string concatenation, and $\mathrm{sim}(\cdot,\cdot)$ is cosine similarity. The filtered candidate set is
\begin{equation}
\mathcal{A}'' = \left\{ a \in \mathcal{A}' \;\middle|\; r(q,a) > \tau \right\} \cup \left\{ a \in \mathcal{A}' \;\middle|\; \mathrm{pattern}(a) \in \mathrm{required\_patterns}(q) \right\}.
\end{equation}

We keep top-50 atomic insights.

\subsection{Multi-Relational Graph Construction}
\label{sec:graph}

We organize the filtered atomic insights into a typed multi-relational graph
\begin{equation}
\mathcal{G} = (\mathcal{V}, \mathcal{E}, \mathcal{R}),
\end{equation}
where nodes $\mathcal{V} = \mathcal{A}''$ represent atomic insights, and $\mathcal{R} = \{\mathrm{Comparative}, \mathrm{Correlative}, \mathrm{Contextual}, \mathrm{Temporal}\}$ denotes four structural relation types. Edges $\mathcal{E}$ are typed and weighted. The graph is implemented as a MultiGraph, allowing multiple parallel edges between the same node pair to capture different relational facets.

\paragraph{Comparative Edge Weight.}
A comparative edge connects two atomic insights if they involve the same metric and aggregation but differ in dimension values. Its weight is the normalized relative difference:
\begin{equation}
w_{\mathrm{comp}}(i,j) = \frac{|v_i - v_j|}{\max(v_i, v_j)},
\end{equation}
where $v_i$ and $v_j$ are the result values. We retain only edges with $w_{\mathrm{comp}}(i,j) > 0.15$.

\paragraph{Correlative Edge Weight.}
A correlative edge connects atomic insights that exhibit statistical dependencies, including CORR-type insights and co-moving TREND-type insights. Its weight is
\begin{equation}
w_{\mathrm{corr}}(i,j) = |r_{ij}| \cdot (1 - p_{ij}),
\end{equation}
where $r_{ij}$ is the Pearson correlation coefficient and $p_{ij}$ is the corresponding $p$-value. Edges with $w_{\mathrm{corr}}(i,j) > 0.6$ are considered strongly correlative.

\paragraph{Contextual Edge Weight.}
A contextual edge connects a global aggregate insight with a local segmented insight. Its weight quantifies the deviation of the local pattern from the global baseline:
\begin{equation}
w_{\mathrm{context}}(i,j) = \max \left( \frac{|s_i - s_j|}{|s_j| + \epsilon}, \; \mathbb{I}[\mathrm{sign}(s_i) \neq \mathrm{sign}(s_j)] \cdot 2 \right),
\end{equation}
where $s_i$ and $s_j$ are the trend slopes or magnitude values of the local and global insights, and $\epsilon$ is a small constant to prevent division by zero.

\paragraph{Temporal Edge Weight.}
A temporal edge connects segmented trend insights of the same metric across adjacent time intervals partitioned by the PELT change-point detection algorithm. Its weight is
\begin{equation}
w_{\mathrm{temp}}(i,j) = |s_i - s_j| \cdot \mathbb{I}[\mathrm{len}(T_i) \geq 3 \wedge \mathrm{len}(T_j) \geq 3],
\end{equation}
where $s_i$ and $s_j$ are the trend slopes of adjacent intervals, and $\mathrm{len}(T)$ is the number of time points in the interval. The minimum segment length of three suppresses over-segmentation.


\subsection{Composite Insight Generation}
\label{sec:composite}

\subsubsection{Definition of Composite Insight}

We extend the composite insight taxonomy to five types:
\begin{equation}
\mathrm{type}(c) \in \{\mathrm{comparative}, \mathrm{correlative}, \mathrm{contextual}, \mathrm{temporal}, \mathrm{causal}\}.
\end{equation}
A composite insight $c$ is a six-tuple:
\begin{equation}
c = (id, \mathrm{type}, \mathrm{atoms}, \mathrm{summary}, \mathrm{confidence}, \mathrm{evidence}),
\end{equation}
where:
\begin{itemize}
    \item $id$: a globally unique identifier;
    \item $\mathrm{type}$: one of the five composite categories above;
    \item $\mathrm{atoms} \subseteq \mathcal{A}''$: the set of constituent atomic insights, with $|\mathrm{atoms}| \geq 2$;
    \item $\mathrm{summary}$: a natural language summary;
    \item $\mathrm{confidence} \in [0,1]$: a confidence score;
    \item $\mathrm{evidence}$: the evidence package containing the SQL queries and results of the supporting atoms.
\end{itemize}
In addition, for every generated composite, we compute a causal score $\mathrm{Causal}(c) \in [0,1]$. If $\mathrm{Causal}(c) > \tau_{\mathrm{causal}}$, the composite is marked as a \emph{causal} composite insight.

\subsection{Community-based Aggregation: Implementation Steps}
\label{subsec:community-aggregation}

\begin{enumerate}
    \item \textbf{Subgraph construction.} For each of the four undirected relation types (Comparative, Correlative, Contextual, Temporal), extract a subgraph \(G_r\) containing only edges of that type. Nodes are atomic insights; edges are the corresponding relations.

    \item \textbf{Community detection.} Run the Leiden algorithm on each subgraph \(G_r\) to obtain communities. Leiden guarantees internal connectivity and works well on weighted graphs.

    \item \textbf{Community filtering.} Keep communities with modularity greater than \(0.3\) and size at least \(3\). Discard communities that are too small or loosely connected.

    \item \textbf{Community summary generation.} For each retained community, call an LLM to generate a higher-order composite insight.
    \begin{quote}
\ttfamily
System: You are a strict reviewer.\\
User: Score the incremental value of the composite insight relative to the atomic insights using the following five criteria:
\begin{enumerate}
    \item \textbf{Novel Synthesis (0--3)}: Does it form a new synthesized conclusion rather than listing atoms?
    \item \textbf{Information Gain (0--2)}: Does it contain new information not directly stated but reasonably derivable from the atoms?
    \item \textbf{Logical Coherence (0--2)}: Is it logically consistent and free of contradictions?
    \item \textbf{Relational Explanation (0--2)}: Does it explain the relational pattern or mechanism among atoms?
    \item \textbf{Decision Utility (0--1)}: Does it provide practical guidance for the analysis goal?
\end{enumerate}
Output the score and a brief reason for each criterion, then give the total score. If total \(\ge 6\), answer ``Yes''; otherwise ``No''.\\
Atomic insights: \{...\}\\
Composite insight: \{...\}
\end{quote}

\end{enumerate}

\subsubsection{Deduplication and Confidence Scoring}

Generated composite candidates undergo two quality-control steps.

\paragraph{Deduplication.}
We deduplicate composites based on the set of constituent atomic insight IDs. If two composites have identical atomic ID sets, only the one with higher confidence is retained.

\paragraph{Confidence Scoring.}
The confidence score $\mathrm{Conf}(c) \in [0,1]$ of a composite insight $c$ is computed as
\begin{equation}
\mathrm{Conf}(c)=\mathrm{consistency}(c)\times \frac{1}{|\mathcal{A}_c|}\sum_{a\in \mathcal{A}_c}\mathrm{sig}(a),
\end{equation}
where \(\mathrm{sig}(a)\in[0,1]\) is the normalized statistical significance of atomic insight \(a\), and \(\mathrm{consistency}(c)\in\{0,1\}\) is a binary logical-coherence check. The field \(\mathrm{evidence}(c)\) records whether the composite is directly validated by a joint SQL query (\texttt{direct}) or indirectly through atomic SQLs (\texttt{indirect}); it does not enter the confidence product. Only composites with \(\mathrm{Conf}(c)>0.6\) are retained. Indirect composites are additionally audited manually in the evaluation.
The causal score $\mathrm{Causal}(c)$ is stored as an additional field, and composites with $\mathrm{Causal}(c) > \tau_{\mathrm{causal}}$ are labeled as causal. This design ensures that the final output is reliable, non-redundant, causally verified, and fully traceable for automated data analysis.

\subsection{Report Generation}
\label{sec:report-generation}

Given the set of composite insights \(\mathcal{C}^*\) and an optional user query \(q\), the report generation module synthesizes an article-level analytical report \(\mathcal{R}\) as follows.

\begin{enumerate}
    \item \textbf{Relevance filtering.} If \(q\) exists, encode \(q\) and the summary of each composite insight \(c\) with an embedding model, and compute cosine similarity \(rel(c,q)\). Set a threshold \(\tau_{\mathrm{rel}}\) and discard composites with \(rel(c,q) < \tau_{\mathrm{rel}}\), keeping only those relevant to the query topic. If \(q\) is absent, keep all composites.

    \item \textbf{Thematic grouping.} Cluster the embeddings of all atomic insights into \(K\) themes (K-means clustering). Assign each selected composite to the theme containing the majority of its atoms. Each theme becomes a section of the report.

    \item \textbf{Narrative generation.} For each section, feed the composites and their evidence (atomic insights, SQL, metadata) into an LLM to generate coherent paragraphs. Ensure all insights are mentioned, transitions are smooth, and every claim is accompanied by evidence references.

    \item \textbf{Report assembly.} Assemble all sections in logical order into a complete report with title, abstract, and conclusion. Output the final report \(\mathcal{R}\), where every claim is traceable to its supporting evidence.
\end{enumerate}

\begin{table*}[t!]
\centering
\small
\caption{Basic Analysis Patterns and SQL Templates}
\label{tab:patterns}
\begin{tabular}{@{}llp{6cm}l@{}}
\toprule
\textbf{Pattern Type} & \textbf{Description} & \textbf{SQL Template Example} \\
\midrule
MAX & Returns maximum value of a metric with associated dimension & \texttt{SELECT MAX(\{metric\}) FROM \{table\}} \\
MIN & Returns minimum value of a metric with associated dimension & \texttt{SELECT MIN(\{metric\}) FROM \{table\}} \\
AVG & Computes the average of a metric & \texttt{SELECT AVG(\{metric\}) FROM \{table\}} \\
SUM & Computes the total of a metric & \texttt{SELECT SUM(\{metric\}) FROM \{table\}} \\
COUNT & Counts the number of records & \texttt{SELECT COUNT(*) FROM \{table\}} \\
GROUP\_BY & Groups by dimension and computes aggregate per group & \texttt{SELECT \{dim\}, \{agg\}(\{metric\}) FROM \{table\} GROUP BY \{dim\}} \\
CORR & Computes correlation between two metrics & \texttt{SELECT CORR(\{m1\}, \{m2\}) FROM \{table\}} \\
TREND & Performs linear regression of metric over time & \texttt{SELECT slope, p\_value FROM (\{table\} time regression)} \\
\bottomrule
\end{tabular}
\end{table*}

\subsection{Basic Analysis Patterns and SQL Templates}
In Table~\ref{tab:patterns}, the basic Analysis Patterns and SQL Templates is shown.

\end{document}